\pdfoutput=1

\documentclass[11pt]{article}

\usepackage[final]{ACL2023}

\usepackage{times}
\usepackage{latexsym}
\usepackage{amsmath}
\usepackage{graphicx}
\graphicspath{{figures/}}
\DeclareGraphicsExtensions{.png,.pdf,.jpg}
\usepackage[T1]{fontenc}
\usepackage[utf8]{inputenc}
\usepackage{microtype}
\usepackage{inconsolata}
\usepackage{booktabs}

\usepackage{arydshln}
\usepackage{multirow}
\usepackage{array}
\usepackage{colortbl}
\usepackage{xcolor}
\usepackage{url}
\usepackage{enumitem}
\usepackage{multicol}
\usepackage{cuted}  

\usepackage{placeins}
\usepackage[most,breakable]{tcolorbox} 
\usepackage{listings}

\newcommand{\datasetname}{\textsc{EarningsInOne}}

\title{Fast Numbers, Slow Language:\\
       Bridging Quantitative and Qualitative Earnings Signals}

\author{
  Ding Yu\textsuperscript{1} \quad
  Zhuo Liu\textsuperscript{1} \quad
  Hao Zhang\textsuperscript{2} \quad
  Hangfeng He\textsuperscript{1} \\[4pt]
  \textsuperscript{1}University of Rochester \quad
  \textsuperscript{2}Rochester Institute of Technology \\
  \texttt{\{ding.yu, zhuo.liu, hangfeng.he\}@rochester.edu} \quad
  \texttt{hzhang@saunders.rit.edu}
}
\definecolor{clrPr}{RGB}{214,39,40}       
\definecolor{clrNxtOpn}{RGB}{44,160,44}   
\definecolor{clrPair}{RGB}{31,119,180}    
\definecolor{clrFstAct}{RGB}{255,127,14}  

\definecolor{clrPos}{RGB}{42,160,140}     
\definecolor{clrNeg}{RGB}{210,120,50}     

\definecolor{clrAll}{RGB}{44,160,44}      
\definecolor{clrNoSP400}{RGB}{214,39,40}  
\definecolor{clrNoSP500}{RGB}{148,103,189}
\definecolor{clrNoSP600}{RGB}{140,86,75}  

\begin{document}
\raggedbottom
\maketitle

\begin{abstract}
Earnings announcements release two types of information sequentially:
\textbf{quantitative surprise} (numeric earnings-per-share (EPS)/revenue versus analyst estimate) arrives first in press releases and financial news,
processed by algorithmic traders within minutes;
\textbf{qualitative language} (management tone, guidance, question-and-answer (Q\&A) credibility) arrives 30--90\,min later in the earnings conference
call transcript (ECT), requiring human interpretation overnight.
Financial economists have studied quantitative surprise for 50 years;
natural language processing (NLP) researchers have studied qualitative ECT signals for a decade.
Despite studying the same event, the two communities used incompatible
frameworks: different targets (return vs.\ volatility), trading setups
(long top decile and short bottom-decile vs.\ trade-all), and metrics (return spread between top and bottom 20\% (Q5$-$Q1) vs.\ mean squared error MSE);
making direct comparison and connection challenging.

We bridge these communities with \datasetname{}, the first corpus
aligning earnings news, ECTs, and intraday and next-day prices across SP\,1500
(broad U.S.\ equity universe, 2022--2025).
Applying unified trading and evaluation tools to both signal types, we
confirm a clean speed separation: \emph{fast numbers, slow language}:
quantitative surprise peaks at announcement and is largely eliminated
by next market open; qualitative ECT sentiment peaks on the next trading day,
real and tradeable, but hidden under prior transcript-based evaluation
that optimised sign-agnostic volatility with pointwise MSE.
\footnote{Our code is publicly available at \url{https://github.com/piqueyd/Fast-Numbers-Slow-Language}}.
\end{abstract}

\section{Introduction}
\label{sec:intro}

Earnings announcements release two types of information sequentially.
\textbf{Quantitative}: the numeric surprise
(e.g.\ ``Apple beat earnings per share (EPS) by \$0.08, reporting \$1.52 vs.\ \$1.44 expected'')%
 arrives first, actionable within minutes.
\textbf{Qualitative}: management tone and guidance
(e.g.\ ``we are cautiously optimistic heading into third quarter (Q3), though macro
headwinds remain a concern'')%
arrives 30--90\,min later in the
earnings conference call, requiring human interpretation overnight.
Despite studying the same event, two research communities,
financial economists and natural language processing (NLP) researchers,
have examined these signals in isolation using incompatible frameworks.

\paragraph{Two gaps we address.}
Financial economists have studied quantitative surprise for over 50
years \citep{ball1968earnings, bernard1989pead, fink2021review}, but focused primarily
on next market open and multi-day horizons.
We use news-flow timestamps to anchor the \emph{first tradable realization}
of numeric information: without precise event-level timestamps,
numeric signal exhausted within minutes is indistinguishable from
language signal maturing overnight \citep{christensen2025warpspeed}. NLP researchers have studied qualitative earnings conference call transcript (ECT) signals \citep{qin-yang-2019-say,
yang2020html, yang2022numhtml} but without a practical trading framework
(no buy/sell selection based on signal confidence,
all stocks equally weighted and traded) and metrics not designed for directional trading
(mean squared error (MSE) over sign-agnostic volatility across all stocks), leaving real trading value unexplored:
a signal is only as useful as the positions it generates, while the financial economics community has long provided
a mature trading framework: long top-decile, short bottom-decile
(buy strongest signals, sell weakest),
and quantitative asset management offers the natural evaluation tools
for signal ranking and strategy evaluation: information coefficient (IC,
Spearman rank correlation between signal and returns) and Sharpe ratio
(a widely used measure of risk-adjusted trading performance: return earned per unit of risk) \citep{grinold2000}.

\paragraph{Our bridge.}
Neither community could see how the other's signal performs,
as incompatible frameworks made direct cross-community comparison challenging,
leaving each community with only part of the picture.
We connect these approaches by asking: \emph{what happens when evaluating both quantitative and qualitative earnings information together,
at announcement resolution and next market open, with intraday holding periods and unified financial evaluation tools?}
To answer this, we introduce \datasetname{}, the first corpus
aligning earnings news, ECTs, and intraday and next-day prices for
the same events across SP\,1500.
Our answer reveals a clean separation by processing speed:
quantitative surprise is \emph{fast}: IC peaks at announcement
and decays to near zero by the next market open;
qualitative sentiment is \emph{slow}: predictive content peaks
on the next trading day.
The bridge yields concrete benefits for both communities: for the financial economics community,
we evaluate quantitative earnings surprises at announcement-level timestamps,
revealing short-horizon alpha that multi-day studies miss
\citep{martineau2022rest}; for the NLP community, applying a long top-decile/short bottom-decile trading framework
and quantitative asset management metrics (IC, Sharpe)
reveals that ECT sentiment is a tradeable signal with real
economic value.

\paragraph{Contributions.}
\begin{enumerate}[leftmargin=*, label=(\arabic*)]

  \item \textbf{Cross-community bridge:}
  We propose a unified evaluation framework connecting financial economics
  and NLP earnings research, sharing a trading convention (long top-decile, short bottom-decile),
  metrics (IC, Sharpe), and data (earnings events),
  enabling direct cross-community comparison.

  \item \textbf{Dataset:}
  \datasetname{} is the first corpus resolving all three
  cross-community pain points simultaneously: precise timestamps from news, intraday and next-day prices, and ECT+earnings news alignment across SP\,1500.

  \item \textbf{Empirical findings:}
  Applying the unified framework reveals two results.
  \textit{Fast numbers, slow language}: quantitative surprise peaks
  at announcement and decays by next market open; qualitative ECT sentiment
  peaks on the next trading day.
  \textit{NLP signals have real economic value}: evaluated under a practical trading framework,
  ECT sentiment produces actionable trading returns previously hidden by
  misaligned metrics (MSE over sign-agnostic volatility targets).

\end{enumerate}

\section{Related Work}
\label{sec:related}

\paragraph{Quantitative Earnings Prediction: The PEAD Literature.}
The PEAD phenomenon---first identified by \citet{ball1968earnings} and
formalized by \citet{bernard1989pead}---documents that stock returns
drift in the direction of earnings surprise for days to weeks after
announcement, measurable via standardized unexpected earnings (SUE)
and evaluated with quintile spread (Q5$-$Q1) and cumulative abnormal
return (CAR) \citep{latane1979standardized}.
Revenue surprises carry incremental predictive content beyond EPS
\citep{JEGADEESH2006147}, and management guidance moves prices
independently at announcement time \citep{billings2015guidance}.
Critically, \citet{martineau2022rest} show that multi-day PEAD has been
non-existent for non-microcap stocks since 2006, prices now fully
reflect earnings surprises on the announcement date itself, as
near-instantaneous price jumps replace gradual drift
\citep{christensen2025warpspeed}.
This compression of the exploitable window motivates our evaluation
at short holding periods: quantitative surprise alpha has not disappeared, it
has accelerated, and this work recovers it by evaluating with
immediate actions rather than waiting for the next market open.

\vspace{-2pt}
\paragraph{Qualitative Earnings Prediction: NLP and ECT-Based Models.}
The dominant NLP paradigm applies multimodal learning over ECT audio
and text \citep{qin-yang-2019-say, yang2020html, cao-etal-2024-ecc, cao2025risklabs},
with extensions for numerical reasoning
\citep{yang2022numhtml, BridgingNumbersandNarratives},
graph-based inter-stock modeling
\citep{sawhney-etal-2020-voltage, Liu_Zhu_Wang_Ma_Yin_Zheng_2024},
and agentic workflows \citep{li-lu-2026-decoding}.
Recent work questions the informativeness of ECT embeddings as a
standalone signal: same-company transcripts are highly formulaic
across periods, limiting incremental information beyond boilerplate
\citep{shih2025company, SS}.
\citet{meursault2023pead} show that text-based ECT patterns produce
drift larger than classic PEAD, providing independent support for our
fast/slow distinction --- yet all these works evaluate at the multi-day
horizon where drift has largely disappeared (see above) under misaligned
pointwise metrics; \S\ref{sec:two-communities} elaborates.
Instead, this work applies LLM sentiment scoring over ECTs and news
under a long/short trading framework.

\section{Framework: Fast Numbers, Slow Language}
\label{sec:framework}

\begin{figure*}[t]
  \centering
  \includegraphics[width=\linewidth]{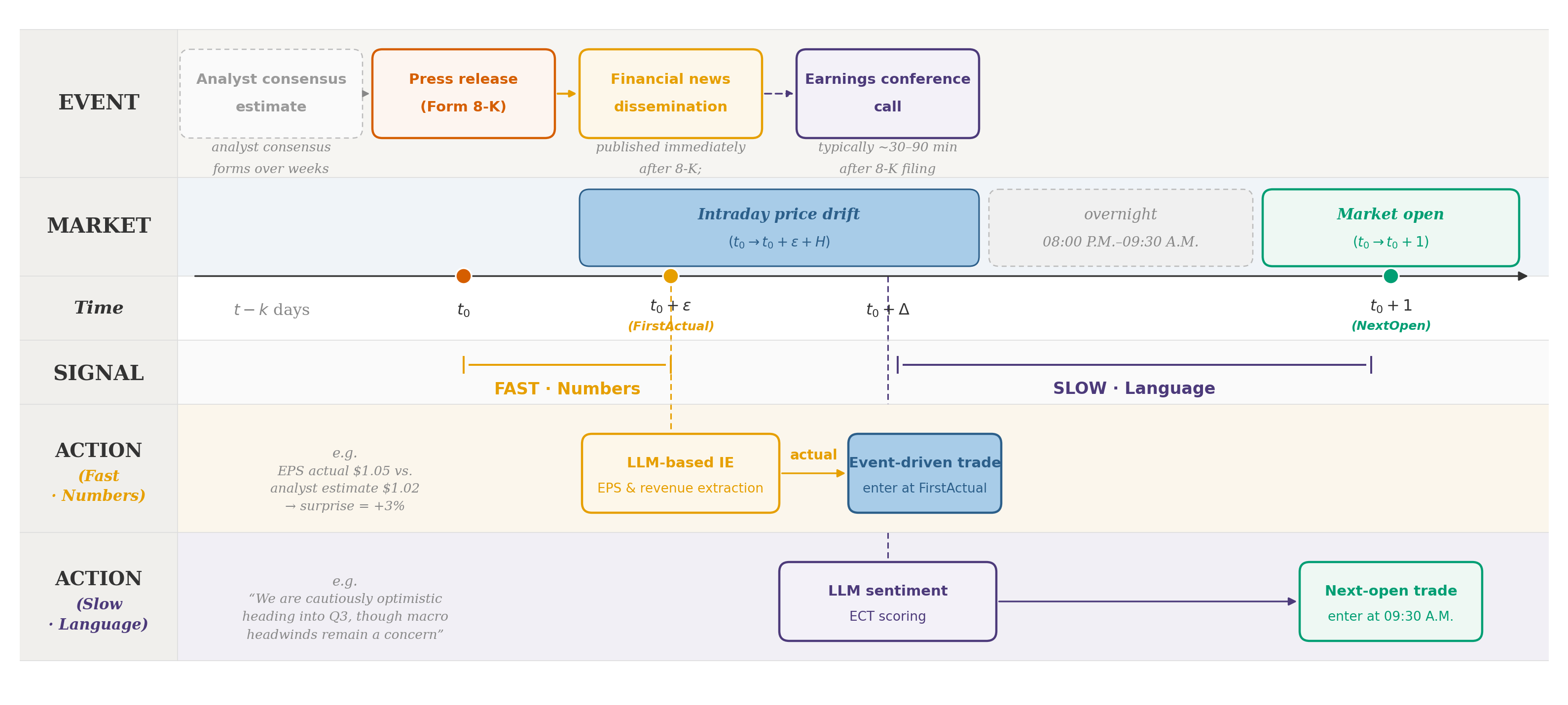}
  \caption{The earnings event timeline and two trading approaches.
  \textbf{EVENT:} Four sequential stages around announcement time $t_0$:
  analyst estimate (weeks prior); press release (Form~8-K, $t_0$)
  discloses \textbf{what} the numbers are; financial news
  ($t_0{+}\varepsilon$, $\varepsilon\approx0$--$5$\,min) contextualises
  actuals against estimates; earnings conference call ($t_0{+}\Delta$,
  $\Delta\approx30$--$90$\,min) conveys \textbf{how} results are framed
  and provides forward guidance.
  \textbf{FAST (Numbers):} LLM-based information extraction yields
  EPS/revenue surprise from news; strategies enter at \textit{FirstActual}
  and hold for up to 60\,min
  (\textit{e.g.\ EPS actual \$1.05 vs.\ analyst estimate \$1.02 $\to$ surprise\,=\,+3\%}).
  \textbf{SLOW (Language):} LLM sentiment scored over the earnings
  conference call transcript; strategies enter at \textit{NextOpen}
  and hold for up to 390\,min (next-day close)
  (\textit{e.g.\ ``We are cautiously optimistic heading into Q3, though macro headwinds remain a concern.''}).}
  \label{fig:comparison}
\end{figure*}

\subsection{The Earnings Event Timeline}
\label{sec:formalize}

We model an earnings event as four sequential stages
(Figure~\ref{fig:comparison}):

\begin{enumerate}[label=(\arabic*), leftmargin=*]
  \item \textbf{Analyst consensus estimate} ($t < t_0$):
  analyst expectations accumulate over weeks via reports and news
  (e.g.\ analyst estimate\,=\,\$0.50).
  \item \textbf{Earnings press release} ($t = t_0$):
  reported actuals disclosed immediately, triggering price discovery
  (e.g.\ EPS actual\,=\,\$0.60).%
  \footnote{Announcements occur either AMC (after-market close, 16:00--20:00 ET)
  or BMO (before-market open, 06:00--09:30 ET), determining the entry window;
  AMC trades at next session open unless entered via extended hours,
  BMO at the regular open. We enter at \textit{FirstActual} in both cases.}
  \item \textbf{Financial news dissemination} ($t = t_0{+}\varepsilon$,
  $\varepsilon \approx 1$--$5$\,min): journalists publish articles co-reporting actuals and estimates
  (\textit{e.g.\ EPS actual \$1.05 vs.\ analyst estimate \$1.02}).
  \item \textbf{Earnings conference call (ECT)} ($t = t_0{+}\Delta$,
  $\Delta\approx30$--$90$\,min): management call with analysts,
  \textbf{primary text source of prior NLP work}, typically available after
  stage~(3).
\end{enumerate}

Stages~(1)--(3) carry \textit{quantitative} information 
(numeric actuals, consensus estimates, and surprise magnitude) 
and are all reflected in financial news articles; 
while stage~(4) carries primarily \textit{qualitative} information recorded as earnings conference call transcripts (ECTs): 
management tone, credibility signals, and forward guidance language.
The two are structurally linked: the ECT is a \textit{linguistic
encoding of the numbers}, available only after those numbers have been
public for 30--90 minutes, with management exercising
\textit{discretionary framing} over how results are presented:
a company can beat EPS while signalling caution, or miss while
projecting confidence,  language and numbers can tell entirely opposite stories.

This creates two consequences: \textit{informational complementarity}
(numbers encode \textbf{what} was reported; language encodes \textbf{how}
management chose to present it) and a speed difference rooted in the
timeline itself: the 8-K precedes the ECT by 30--90 min, making numeric
surprise actionable before qualitative language exists; and numbers are
unambiguous and machine-readable; algorithms act simultaneously,
driving rapid price discovery, while tone requires interpretation and
credibility assessment, so market consensus on qualitative signals forms
usually overnight.

\begin{table*}[!ht]
\centering
\small
\setlength{\tabcolsep}{7pt}
\renewcommand{\arraystretch}{1.35}

\begin{tabular}{%
  >{\raggedright\arraybackslash}p{2.0cm}%
  >{\raggedright\arraybackslash}p{3.5cm}%
  >{\raggedright\arraybackslash}p{4.0cm}%
  >{\columncolor{gray!10}\raggedright\arraybackslash}p{4.0cm}%
}
\toprule
\textbf{Dimension}
& \textbf{Financial Economics Community}
& \textbf{NLP Community}
& \textbf{This Work (Bridge)} \\
\midrule

\textbf{Target}
  & Post-ann.\ return$^{[1\text{-}5]}$
  & Post-ann. volatility$^{[6\text{-}11]}$ \newline / Post-ann. return direction $^{[12]}$
  & \cellcolor{gray!10}Post-ann.\ return \\
\specialrule{0.3pt}{2pt}{2pt}

\textbf{Data \&} \newline \textbf{Signal}
  & EPS/revenue surprise
  \newline $\Delta$EPS\% + SUE$^{[1\text{-}5]}$
  & ECT (text / audio); \newline language embeddings$^{[6\text{-}12]}$
  & \cellcolor{gray!10}Earnings news + ECT; \newline $\Delta$EPS\% + Sentiment socre \\
\specialrule{0.3pt}{2pt}{2pt}

\textbf{Entry \& Horizon}
  & Next market open; \newline 3--30 days$^{[1\text{-}5]}$
  & $\tau$-day post call; \newline 3--30 days$^{[6\text{-}12]}$
  & \cellcolor{gray!10}At announcement (5--60\,min); \newline next market open (5--390\,min) \\
\specialrule{0.3pt}{2pt}{2pt}

\textbf{Trading} \newline \textbf{Mechanism}
  & Long top-decile \newline short bottom-decile$^{[1\text{-}5]}$ \newline (cross-sectional sort)
  & None (implicitly trade all)$^{[6\text{-}12]}$
  & \cellcolor{gray!10}Long top-decile \newline short bottom-decile \newline (historical percentile) \\
\specialrule{0.3pt}{2pt}{2pt}

\textbf{Metric}
  & Q5$-$Q1 / CAR$^{[1\text{-}5]}$
  & MSE$^{[6\text{-}11]}$ / F1$^{[12]}$
  & \cellcolor{gray!10}Q5$-$Q1, IC, SR, Tot.\ Ret. \\

\bottomrule
\end{tabular}

\caption{Methodological comparison across the financial economics and NLP
communities and this work.
\textbf{Financial economics:}
[1]~\citet{ball1968earnings};
[2]~\citet{bernard1989pead};
[3]~\citet{latane1979standardized};
[4]~\citet{martineau2022rest};
[5]~\citet{benrephael2024mindthegap}.
\textbf{NLP:}
[6]~\citet{qin-yang-2019-say};
[7]~\citet{yang2020html};
[8]~\citet{yang2022numhtml};
[9]~\citet{sawhney-etal-2020-voltage};
[10]~\citet{cao-etal-2024-ecc};
[11]~\citet{SS};
[12]~\citet{Liu_Zhu_Wang_Ma_Yin_Zheng_2024}.
Post-ann.~=~post-announcement (i.e.\ after the earnings release); Post-announcement volatility~=~log rolling standard deviation of daily returns \citep{kogan2009predicting} ($\tau \in \{3,7,15,30\}$ days);
SUE~=~$(\mathrm{EPS}_{\mathrm{act}} - \mathrm{EPS}_{\mathrm{est}}) / \sigma(\varepsilon)$ (standardised unexpected earnings),
where $\sigma(\varepsilon)$ is the standard deviation of historical analyst forecast errors;
$\tau$-day~=~$\tau \in \{3,7,15,30\}$ days post earnings conference call;
MSE~=~mean squared error; F1~=~harmonic mean of precision and recall;
IC~=~information coefficient (Spearman rank); Q5$-$Q1~=~return spread between top and bottom quintile;
SR~=~Sharpe ratio; Tot.\ Ret.~=~total return (equivalent to CAR).
long top-decile / short bottom-decile~=~buy stocks with strongest signals, sell stocks with weakest signals (top and bottom 10\%);
cross-sectional sort~=~stocks ranked against peers within the same session;
historical percentile~=~stocks ranked against the market-wide historical score distribution.}
\label{tab:community-comparison}
\end{table*}

\subsection{One Phenomenon, Two Communities}
\label{sec:two-communities}

Earnings announcements have been studied through two lenses:
the \emph{financial economics community} focuses on numeric surprise
and directional return; the \emph{NLP community} focuses on ECT-based
verbal cues (management tone, guidance language, and Q\&A credibility)
as incremental predictors beyond the headline numbers.
Table~\ref{tab:community-comparison} summarises how these beliefs
translate into five methodological divergences (see Appendix~\ref{sec:appendix-community}
for full details).

\begin{itemize}[leftmargin=1.5em, itemsep=0pt]
  \item \textbf{Target:} financial economics predicts directional
  return; NLP predominantly predicts sign-agnostic volatility;
  we target directional return, the most actionable measure of market reaction to earnings.
  \item \textbf{Data \& Signal:} financial economics uses vendor
  numeric surprise ($\Delta$EPS\%, SUE); NLP uses ECT language
  embeddings; this work combines both via news-based information extraction (IE) and LLM
  sentiment (\S\ref{sec:method}).
  \item \textbf{Entry \& Horizon:} both communities enter at
  next market open and hold for 3--30 days, missing the intraday window
  where fast surprise alpha concentrates \citep{martineau2022rest};
  we enter at \textit{FirstActual} (5--60\,min) and
  \textit{NextOpen} (5--390\,min), each at its natural speed.
\end{itemize}

The fourth and fifth divergences, \textbf{trading framework and
evaluation metric}, are causally linked: the absence of a trading
framework in NLP work leads directly to a misaligned metric.
The financial economics community goes long the top decile, short the
bottom decile \citep{bernard1989pead}, treating earnings as a
cross-sectional ranking problem naturally evaluated by IC and Sharpe.
The NLP community evaluates pointwise with MSE over all announcing
stocks, implicitly treating every event as equally tradeable.
\textit{Consider a session with 10 announcing stocks: a financial
economist goes long the top-1 beat and short the worst-1 miss,
measuring whether the signal correctly separates them (IC); an NLP
model predicts all 10 and optimises MSE, indifferent to whether the
high-conviction predictions are correct.}
We adopt IC and Sharpe as primary metrics and retain Q5$-$Q1 for
comparability with the PEAD literature.

\section{Dataset: \datasetname{}}
\label{sec:dataset}

\subsection{Novelty and Positioning}

\datasetname{} is the first dataset enabling unified evaluation of
fast quantitative and slow qualitative earnings signals.
No prior dataset simultaneously covers both dimensions:

\begin{itemize}
  \item \textbf{Intraday and interday price resolution.}
  Both communities evaluate at next market open; \datasetname{} pairs each
  event with minute-level price bars, enabling intraday evaluation of
  fast signals and next market open evaluation of slow signals.

    \item \textbf{Earnings news: precise timestamps, real-time actuals, and complementary narrative.} Prior NLP work relies solely on ECTs \citep{ectsum, SS}; financial news articles additionally provide precise publication timestamps enabling intraday entry,
real-time EPS actuals without vendor lag \citep{hoechle2015time},
and a complementary qualitative narrative that pairs with ECTs.
\end{itemize}

\subsection{Coverage, Split, and Availability}
\datasetname{} covers the SP\,1500 universe, comprising SP\,500
(large-cap), SP\,400 (mid-cap), and SP\,600 (small-cap)\footnote{SP\,600 represents only $\sim$100 events after data cleaning, 
reflecting lower financial news coverage for smaller firms; 
results are thus primarily driven by large- and mid-cap stocks.}, spanning
2022--2025.
The corpus comprises 5{,}428 earnings events across 701 unique
tickers, covered by 257{,}815 news articles.
News articles, intraday prices, and analyst consensus estimates are all sourced from data vendor EODHD Historical Data\footnote{Data vendor: \href{https://eodhd.com/}{EODHD Historical Data}}.
We plan to release verified article URLs, earnings event metadata,
and the alignment schema upon publication; full dataset details are in
Appendix~\ref{sec:appendix-dataset}.

\section{Unified Evaluation Framework}
\label{sec:method}

\subsection{Signal Extraction}
\label{sec:llm-ie}

We apply GPT-5-mini\footnote{\href{https://developers.openai.com/api/docs/models/gpt-5-mini}{GPT-5-mini Model Card}. Knowledge cutoff: May 31, 2024, slightly overlapping our test period (2024--2025).} to earnings-related news article to extract EPS
actuals and revenue actuals; estimates are sourced from EODHD Historical Data pre-release.%
\footnote{We use vendor estimates (EODHD Historical Data) since news estimates
do not always co-appear with actuals and may add delay;
EODHD pre-release estimates are available 1--3 days before
the press release at negligible cost.}
We denote \textbf{(EN)} signals as LLM-extracted actuals with
EODHD pre-release estimates; \textbf{(EOD)} as vendor-confirmed
EODHD actuals and estimates.
For qualitative signals, we apply GPT-5-mini to each ECT or news
article to produce a scalar sentiment score in $[-1, 1]$ reflecting
overall management tone and earnings narrative.
Full prompt in Appendix~\ref{sec:appendix-prompt}.

\subsection{Trading Mechanisms}
\label{sec:signal}

Both strategy types go long top 10\% and short bottom 10\% signals
within each session, following canonical trading settings.
They differ in how rankings are formed:
\begin{itemize}[leftmargin=1.5em]
  \item \textbf{Canonical PEAD} \citep{bernard1989pead}:
  contemporaneous cross-sectional sort within each session;
  applicable at \textit{NextOpen} when all announcements are known
  (e.g.\ 10 stocks report the same day: the top beat is bought, the worst miss sold).
  \item \textbf{Historical percentile:}
  ranks each signal against an expanding-window cumulative distribution function (CDF) of the past 12 months.
  Required for \textit{FirstActual} entry, as stocks announce
  throughout the session, making contemporaneous sorting infeasible
  (e.g.\ a stock scoring in the top 10\% of the whole market's past 12-month signal distribution is bought).
\end{itemize}
Full trading setting in Appendix~\ref{sec:appendix-strategies}.

\subsection{Entry, Horizon, and Return Measurement}
\label{sec:returns}

We define two entry anchors:
\begin{itemize}[leftmargin=1.5em]
  \item \textbf{FirstActual}: earliest article with confirmed EPS/revenue
  actual; estimate from EODHD; intraday horizons $H \in \{5,15,30,60\}$\,min.%
  \footnote{Typically $\approx$07:00 for BMO announcements and
  $\approx$16:00--16:30 for AMC announcements; \textit{FirstActual} horizons are capped at 60\,min since AMC occur after market close, leaving no extended intraday window.
  full details in Appendix~\ref{sec:appendix-timestamps}.}
  \item \textbf{NextOpen}: 09:30\,am open on the following day,
  standard entry anchor in both communities; intraday horizons
  $H \in \{5,15,30,60,120,180,390\}$\,min.%
  \footnote{390\,min = end of regular trading session.}
\end{itemize}

For each anchor $\tau$ and horizon $H$, log return:
\[
  r_{i,\tau,H} = \log P^{\text{close}}_{i,\tau^++H} -
  \log P^{\text{open}}_{i,\tau^+}
\]
where $i$ = stock, $P$ = price, $\tau^+$ is the first $H$-min bar starting at least 1\,min after
$\tau$ (1-min information processing lag); e.g.\ $\tau$=16:00:32 $\to$ $\tau^+$=16:02 $\to$ $\tau^+$+$H$=16:07 (if $H$=5\,min).

\section{Experiments}
\label{sec:experiments}

\subsection{Experimental Setup}
\label{sec:exp-setup}
\label{sec:results}

\paragraph{Setup.}
We evaluate on the SP\,1500 universe using a strict temporal split:
train 2022, validation 2023, test 2024--2025.
For each strategy the best horizon is selected on the
validation set and held fixed for the test.
Full details in Appendix~\ref{sec:appendix-strategies}.

\paragraph{Evaluation Metrics.}
\label{sec:eval-metrics}
We report five metrics; full definitions in
Appendix~\ref{sec:appendix-metrics}:
\begin{itemize}[leftmargin=1.5em, itemsep=0pt,]
  \item \textbf{IC ($t_{\mathrm{NW}}$)}: Spearman rank IC
  \citep{grinold2000} with Newey-West $t$-statistic; measures
  pure signal quality; position-agnostic.
  \item \textbf{Q5$-$Q1 bp ($t$)}: Fama-MacBeth session-averaged
  quintile spread \citep{bernard1989pead} in basis points; adds
  selectivity to IC, model-free and matching PEAD reporting.
  \item \textbf{Sharpe}: annualised risk-adjusted return;
  measures overall strategy quality by incorporating signal quality, selectivity, and sizing;
  the most complete yet implementation-sensitive.
  \item \textbf{Tot.\ Ret.}: cumulative total return.
  \item \textbf{$N$}: number of trades.
\end{itemize}

\paragraph{Passive, Na\"{i}ve, and Prior-NLP Baselines.}
Market benchmarks with no earnings signal:
\textbf{SPY} and \textbf{QQQ}\footnote{SPY and QQQ are the most widely traded passive ETFs tracking the S\&P 500 and Nasdaq-100; \href{https://finance.yahoo.com/quote/SPY/}{SPY}, \href{https://finance.yahoo.com/quote/QQQ/}{QQQ}.} (S\&P 500 and Nasdaq-100 index funds);
\textit{Random L/S} and \textit{Long-only} (no-signal floors).
We additionally include two prior-NLP baselines:
\textit{Ticker-Mean} (mean of historical same-ticker post-announcement returns)
and \textit{Embedding}\footnote{\href{https://platform.openai.com/docs/guides/embeddings}{text-embedding-3-large (OpenAI)}.} (ECT embeddings passed through a 2-layer MLP).
\citet{SS} show that ECT embeddings exhibit stronger within-ticker
than cross-ticker similarity,%
\footnote{We confirm this in \datasetname{}: 0.80 vs.\ 0.67 avg.}
such that embedding predictions largely reflect ticker identity rather than earnings content.

\paragraph{Quantitative Conditions.}
Five variants isolate four factors:
\begin{itemize}[leftmargin=1.5em, itemsep=0pt]
  \item \textbf{Ranking Mechanism} (PEAD vs.\ historical percentile):
  \textit{PEAD-EPS(EN)} uses contemporaneous sort \citep{bernard1989pead};
  \textit{EPS(EN)} uses historical percentile; both at NextOpen,
  isolating the mechanism effect.
  \item \textbf{Entry anchor} (NextOpen vs.\ FirstActual):
  \textit{EPS(EN)} at both anchors isolates entry timing.
  \item \textbf{Source breadth} (EPS vs.\ Joint):
  \textit{Joint(EN)} adds revenue surprise \citep{JEGADEESH2006147} to EPS at FirstActual,
  isolating the revenue contribution.
  \item \textbf{Data quality} (EOD vs.\ EN):%
  \footnote{EODHD actuals have reporting lag and are not live-deployable; \textit{Joint(EOD)} is a data-quality ceiling.}
  vendor-confirmed EODHD actuals vs.\ LLM-extracted,
  providing a data-quality ceiling.
\end{itemize}
\paragraph{Qualitative Conditions.}
Text-based signals evaluated at next market open:
\begin{itemize}[leftmargin=1.5em]
  \item \textit{Sentiment(ECT)} at \textit{NextOpen}: LLM sentiment
  over full ECT transcript; primary prior-work comparison point.
  Two variants explicitly examine the role of numeric information:
  \textit{Sentiment(ECT-mask)} replaces all numbers with \texttt{[NUM]}
  (isolates pure verbal signal) and
  \textit{Sentiment(ECT-informed)} prepends the EPS/revenue surprise
  header (grounding in known beat/miss); both in the main table.
  \item \textit{Sentiment(EN-agg)} at \textit{NextOpen}: sentiment
  aggregated over up to 10 post-earnings news articles up to market open;
  direct comparison with ECT on the same timing window.
\end{itemize}
\subsection{Results}
\label{sec:exp-results}

\begin{figure*}[!ht]
  \centering
  \includegraphics[width=\linewidth]{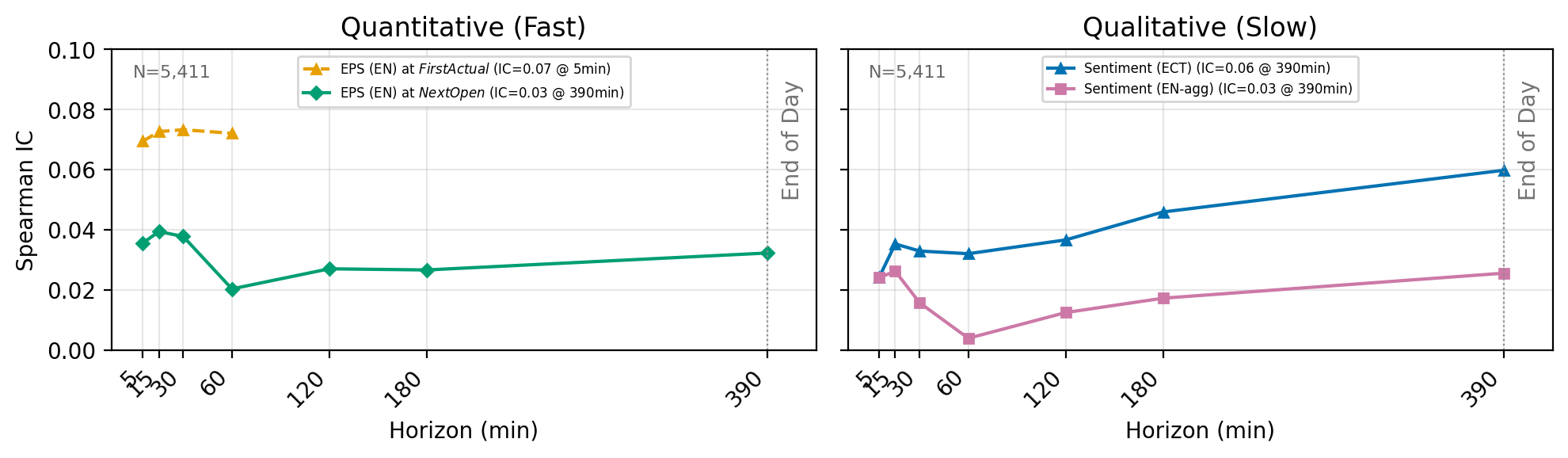}
  \captionsetup{skip=3pt} 
  \caption{Pooled Spearman IC by horizon, SP\,1500 (2022--2025; stable across splits).
  \textbf{Left (Quantitative):} \textit{EPS(EN)} at \textit{FirstActual}
  (dashed) holds IC\,$\approx$\,0.07 stably across 5--60\,min;
  at \textit{NextOpen} (solid) decays from 0.04 to 0.03,
  short-horizon alpha arbitraged away by open.
  \textbf{Right (Qualitative):} \textit{Sentiment(ECT)} rises to
  IC\,=\,0.06 at next-day close; \textit{Sentiment(EN-agg)} dips
  near zero at 60\,min before recovering.
  Horizons beyond 390\,min excluded (IC\,$<$\,0).}
  \label{fig:return-ic}
\end{figure*}

\begin{table*}[!ht]
\centering
\small

\resizebox{\textwidth}{!}{%
\begin{tabular}{ll l ccccc}
\toprule
\textbf{Category} & \textbf{Strategy} & \textbf{Entry anchor} & IC ($t_{\mathrm{NW}}$) & Q5$-$Q1 bp ($t$) & Sharpe & Tot.\ Ret. & $N$ \\
\midrule
\multirow{4}{*}{Na\"{i}ve}
 & Random L/S             & NextOpen    & ---                   & ---               & $-$0.88 & $-$10.5\% & 2426 \\
 & Long-only     & NextOpen    & ---                   & ---               & 0.11    & 5.8\%     & 2077 \\
 & \textit{Ticker-Mean}   & NextOpen    & ---                   & ---               & $-$3.13 & $-$2.2\%  & 2360 \\
 & \textit{Embedding} & NextOpen    & ---                   & ---               & 0.80    & 23.3\%   & 2468 \\
\midrule
\multirow{5}{*}{\shortstack{Quantitative\\(Fast)}}
 & \textit{PEAD-EPS(EN)}  & NextOpen    & 0.03 (0.97)           & 33.9 (1.36)       & \textbf{3.31}    & 29.1\%  & 380 \\
 & \textit{EPS(EN)}    & NextOpen    & $-$0.02 ($-$0.56)     & $-$5.4 ($-$0.12)  & $-$0.31 & $-$4.2\%  & 400 \\
 & \textit{EPS(EN)}    & FirstActual & 0.03 (0.94)           & \textbf{137.7 (3.54)} & 1.81    & 50.4\%  & 400 \\
 & \textit{Joint(EN)}  & FirstActual & 0.05 (1.58)           & 61.3 (1.94)       & 1.79    & 39.9\%  & 296 \\
 & \textit{Joint(EOD)} & FirstActual & \textbf{0.08 (2.40)} & 118.2 (4.13) & 2.43 & \textbf{64.5\%} & 343 \\
\midrule
\multirow{4}{*}{\shortstack{Qualitative\\(Slow)}}
 & \textit{Sentiment(ECT)}          & NextOpen    & \textbf{0.11 (4.04)} & \textbf{153.0 (4.30)} & 2.28 & 63.6\%  & 591 \\
 & \textit{Sentiment(ECT-mask)}     & NextOpen    & 0.07 (2.62) & 118.7 (3.96) & 0.40  & 7.0\%   & 412 \\
 & \textit{Sentiment(ECT-informed)} & NextOpen    & 0.06 (2.10) & 94.1 (2.62)  & \textbf{2.65}  & \textbf{72.0\%}  & 599 \\
 & \textit{Sentiment(EN-agg)}       & NextOpen    & 0.05 (1.89) & $-$4.6 ($-$0.24) & 0.42  & 5.9\%   & 496 \\
\bottomrule
\end{tabular}%
}

\caption{Out-of-sample trading performance, test set (2024--2025 with 2{,}468 events), SP\,1500.
\textbf{Entry anchors:}
\textit{FirstActual} = first news article with confirmed EPS/revenue actuals;
\textit{NextOpen} = 09:30\,am on the following trading day.
\textbf{Metrics:}
IC\,($t_{\mathrm{NW}}$) = Spearman rank IC \citep{grinold2000} computed per event date
($t_{\mathrm{NW}}$: Newey-West $t$-statistic correcting for serial correlation);
Q5$-$Q1\,bp\,($t$) = Fama-MacBeth date-averaged quintile spread \citep{bernard1989pead}
($t$: Fama-MacBeth standard errors);
$|t|>2$ corresponds to $p<0.05$ for both;
$N$ = number of trades.
\textbf{Signals:}
\textit{Joint} = EPS + revenue surprise combined;
EOD: EODHD pre-release estimates; professional-grade vendor actuals with superior data quality, but not available in real time for live deployment;
\textbf{Metric Interpretation:}
IC measures overall signal quality (does the signal rank stocks correctly?);
Q5$-$Q1 measures tail separability (do the extreme quintiles trade well?);
Sharpe measures full strategy performance (does it make money risk-adjusted?).
High IC with low Sharpe indicates the signal ranks stocks well but individual trade returns are noisy and volatile;
high Sharpe with low IC signals selectivity-driven returns (strong performance concentrated in a small number of high-conviction trades).
\textbf{Bold} = best per column within group.
\textbf{Passive Market benchmarks:} SPY and QQQ are the most widely traded ETFs tracking the S\&P\,500 and Nasdaq-100, serving as buy-and-hold baselines: SPY total return 48\% (Sharpe 1.02); QQQ total return 54\% (Sharpe 0.94).}
\vspace{-1em}
\label{tab:trading-results}
\end{table*}

Figure~\ref{fig:return-ic} characterises signal quality across horizons
before we report trading results (Table~\ref{tab:trading-results});
the full table with all variants is in Appendix~\ref{sec:appendix-full-results}.

\paragraph{Baselines.}
Na\"{i}ve baselines yield no reliable alpha: \textit{Random L/S} (Sharpe\,=\,$-$0.88),
\textit{Long-only} (Sharpe\,=\,0.11), \textit{Ticker-Mean} (Sharpe\,=\,$-$3.13).
\textit{Embedding} sign variant (Sharpe\,=\,0.80) outperforms these but
underperforms all signal-based strategies; the magnitude variant is weaker
(full results in Appendix~\ref{sec:appendix-full-results}).

\paragraph{Quantitative Signals.}
\textbf{Mechanism.}
\textit{PEAD-EPS(EN)}%
\footnote{\textit{PEAD-Joint(EN)} does not improve over \textit{PEAD-EPS(EN)}; full results in Appendix~\ref{sec:appendix-full-results}.}
achieves Sharpe\,=\,3.31 despite IC\,=\,0.03 ($t$\,=\,0.97);
high Sharpe reflects extreme selectivity (top and bottom 10\% long/short), not broad ranking quality.

\textbf{Entry anchor.}
\textit{EPS(EN)} at \textit{FirstActual} (IC\,=\,0.03, Sharpe\,=\,1.81, Q5$-$Q1\,=\,137.7\,bp) substantially exceeds the \textit{NextOpen} variant (IC\,=\,$-$0.02, Sharpe\,=\,$-$0.31), confirming intraday alpha is fully arbitraged away by open.

\textbf{Source breadth.}
\textit{Joint(EN)} raises IC to 0.05 over \textit{EPS(EN)} but reduces Q5$-$Q1 and Sharpe, suggesting revenue surprise improves cross-sectional ranking but reduces tail concentration.

\textbf{Data quality.}
\textit{Joint(EOD)} achieves IC\,=\,0.08 ($t$\,=\,2.40), Sharpe\,=\,2.43, outperforming \textit{Joint(EN)};
data quality matters at the margin.%
\footnote{Full EN vs.\ EODHD actuals validation in Appendix~\ref{sec:appendix-validation}.}

\paragraph{Qualitative Signals.}
\textbf{ECT sentiment (vanilla).}
\textit{Sentiment(ECT)} achieves IC\,=\,0.11 ($t$\,=\,4.04),
Sharpe\,=\,2.28, Q5$-$Q1\,=\,153.0\,bp ($t$\,=\,4.30),
all three metrics stronger and more significant than any
quantitative variant at \textit{NextOpen}, confirming ECT
sentiment is a tradeable signal at \textit{NextOpen}.

\textbf{ECT ablations.}
\textit{Sentiment(ECT-mask)} achieves IC\,=\,0.07 but Sharpe\,=\,0.40;
the masking disrupts score distributions for position sizing,
producing noisier returns.
\textit{Sentiment(ECT-informed)} reverses this: Sharpe\,=\,2.65
(highest overall) but IC\,=\,0.06; numeric grounding
boosts tail concentration without improving average ranking.%
\footnote{All three ECT variants show similar pooled IC
and high pooled Spearman rank correlations (0.82/0.80),
suggesting overlapping information despite performance differences.}

\textbf{News sentiment.}
\textit{Sentiment(EN-agg)} (IC\,=\,0.05, Sharpe\,=\,0.42) underperforms ECT sentiment across all metrics,
suggesting news articles carry less qualitative signal than conference calls.

\paragraph{Fast vs.\ Slow: The Full Picture.}
Mirroring Figure~\ref{fig:return-ic}: numeric surprise peaks at \textit{FirstActual}
(IC\,=\,0.03, Sharpe\,=\,1.81) and decays by \textit{NextOpen};
ECT sentiment peaks at \textit{NextOpen} (IC\,=\,0.11, Sharpe\,=\,2.28).
Fast numbers is arbitraged away; slow language persists overnight.

\subsection{Analysis}
\label{sec:exp-analysis}

\paragraph{Why fast numbers, slow language?}
Numeric surprise is a single scalar actionable by algorithms within minutes; language requires overnight interpretation before institutional investors can deploy their large capital pools.

\paragraph{What does LLM sentiment capture?}
Decomposing ECT into 23 fine-grained dimensions
(Appendix~\ref{sec:appendix-decompose}), three patterns emerge:
backward-looking financial quality dominates (\textit{cash flow} IC\,=\,0.055
vs.\ \textit{guidance direction} 0.020); analyst reception is informative
(\textit{analyst surprise/tone} IC\,=\,0.037--0.038, reflecting real-time
expert interpretation); and \textit{surprise novelty} is negatively predictive
(IC\,=\,$-$0.029), as novel disclosures create disagreement over direction.

\begin{table}[h]
\centering
\small
\begin{tabular}{l ccc}
\toprule
& \textit{EPS(EN)} & \textit{Sent(ECT)} & \textit{Sent(EN-agg)} \\
\midrule
\textit{EPS(EN)}       & 1.00    & 0.21    & 0.38    \\
\textit{Sent(ECT)}     & 0.21    & 1.00    & 0.52    \\
\textit{Sent(EN-agg)}  & 0.38    & 0.52    & 1.00    \\
\bottomrule
\end{tabular}

\caption{Spearman rank correlations (2022--2025, $N{\approx}5{,}000$; stable across splits).}
\label{tab:signal-corr}

\end{table}
\vspace{-1em}

\paragraph{Cross-signal correlations.}
The signals exhibit a nuanced correlation structure (Table~\ref{tab:signal-corr}).
\textit{Sentiment(ECT)} is largely orthogonal to \textit{EPS(EN)}
surprise ($\rho$\,=\,0.21), confirming that ECT sentiment captures management tone and framing, not just the headline beat-or-miss.
\textit{Sentiment(EN-agg)} correlates more strongly with \textit{EPS(EN)}
($\rho$\,=\,0.38), as expected: post-announcement news articles co-report
actuals alongside narrative, partially echoing the numeric outcome.
\textit{Sentiment(ECT)} and \textit{Sentiment(EN-agg)} are moderately correlated
($\rho$\,=\,0.52), both covering the same event.

\paragraph{Combining numbers and language.}
The two signals peak at incompatible horizons: \textit{EPS(EN)} peaks at \textit{FirstActual}
but decays by \textit{NextOpen}, where \textit{Sentiment(ECT)} peaks.
\textit{Sentiment(ECT-informed)} trades off IC for Sharpe (IC\,=\,0.06 vs.\ 0.11; Sharpe\,=\,2.65 vs.\ 2.28),
suggesting ECT already encodes the beat/miss outcome implicitly.
Directly merging \textit{Joint(EN)} or \textit{EPS(EN)} with \textit{Sentiment(ECT)}
at \textit{NextOpen} yields further drops in both metrics relative to \textit{Sentiment(ECT)} (Appendix~\ref{sec:appendix-full-results}).
Using ECT sentiment to extend or exit intraday positions opened at \textit{FirstActual}, dynamically bridging the fast/slow gap, is left to future work.

\paragraph{Selectivity and Coverage.}
Table~\ref{tab:selectivity} shows Sharpe rises as $q$ decreases;%
\footnote{\textit{EPS(EN)} at 5\,min; \textit{Sentiment(ECT)} at 390\,min.}
NLP evaluation implicitly uses $q{=}0.50$ (trade all events),
the worst operating point; we use $q{=}0.10$ in our experiments.

\begin{table}[h]
\centering\small
\begin{tabular}{lccccc}
\toprule
$q$ & 0.50 & 0.40 & 0.30 & 0.20 & 0.10 \\
\midrule
\textit{Sent(ECT)} & 0.49 & 0.60 & 0.98 & 1.42 & 2.32 \\
\textit{EPS(EN)}   & 1.53 & 1.23 & 1.57 & 2.21 & 2.27 \\
\bottomrule
\end{tabular}
\caption{Sharpe by quantile threshold $q$ ($q$ = top/bottom fraction traded),
test set (2024--2025).}
\label{tab:selectivity}
\end{table}
\vspace{-1em}

\paragraph{Model robustness.}
Claude Haiku 4.5 confirms the signal at IC\,=\,0.08 ($t$\,=\,2.43);
Gemini 3 Flash confirms the signal at IC\,=\,0.08 ($t$\,=\,2.77),
confirming robustness across LLM families \footnote{\href{https://www.anthropic.com/claude/haiku}{Claude Haiku 4.5 Model} and \href{https://ai.google.dev/gemini-api/docs/models/gemini-3-flash-preview}{Gemini 3 Flash Model} }.

\paragraph{Transaction costs.}
After 1\,bp one-way per trade, \textit{EPS(EN)} and \textit{Sentiment(ECT)}
retain substantial total returns of 42.4\% and 51.8\%, with further reduction possible via execution optimisation.

\section{Conclusion}
\label{sec:conclusion}
Two communities have studied earnings announcements for decades with incompatible frameworks.
We bridge them through \datasetname{} and a unified evaluation framework,
revealing a clean speed separation: numeric surprise peaks at announcement;
ECT sentiment peaks at next-open, largely orthogonal.
Both were capturing real but complementary phenomena, qualitative signals yield real trading returns when evaluated correctly.
More broadly, our findings suggest that language signals may show better value
when evaluated at their natural speed, a challenge shared by data-rich domains
mixing structured and textual signals arriving asynchronously,
such as healthcare (vital signs vs.\ clinical notes).

\section*{Limitations}
\textbf{Limitations.}
\textit{Data:} licensing restricts public release of raw news and price data;
we release metadata, IE annotations, and the alignment schema.
\textit{Scope:} experiments cover U.S.\ equity markets only;
generalisation to other markets or asset classes is left to future work.
\textit{Backtesting:} results assume no transaction costs and perfect fill
at 5-min bar open; strategies use canonical equal-weight sizing
without position limits or risk controls, which may understate or overstate live performance.
\textit{LLM extraction:} systematic biases on financial jargon or low-coverage tickers
are the primary failure mode ($<$2\% of events, typically unit-scaling mismatches
such as per-share vs.\ aggregate revenue);
sentiment scoring uses a single-pass prompt without advanced prompt engineering
(e.g.\ chain-of-thought, self-consistency), leaving room for improvement.
\textit{Stock relations:} inter-stock dependencies (e.g.\ sector contagion,
supply-chain linkages) are not modelled; signals are treated independently per event. \textit{Risks:} the trading signals described are intended for research purposes;
deployment in live markets without proper risk management could result in financial losses.
The intraday strategies may exacerbate short-term price volatility around earnings announcements.
LLM extraction errors, if undetected, could generate misleading signals in automated pipelines.


\section*{Ethics Statement}
All news and price data were obtained through licensed commercial APIs.
No material non-public information was used.
Trading results are academic backtests and do not constitute investment
advice.

\bibliography{anthology,custom}

\newpage
\appendix

\section{Detailed Community Comparison}
\label{sec:appendix-community}

The following elaborates on each dimension of Table~\ref{tab:community-comparison}
with a paragraph-level comparison across the financial economics community,
the NLP community, and this work.

\paragraph{Target.}
\begin{itemize}[leftmargin=1.5em]
  \item \textbf{Financial economics community:} post-announcement return
  (the directional price move following the earnings release).
  \item \textbf{NLP community:} predominantly post-announcement volatility
  (the log rolling standard deviation of daily returns),
  which is sign-agnostic by construction; a minority of works
  target return directly.
  \item \textbf{This work (bridge):} post-announcement return,
  the most direct market reaction to earnings, fundamental to both
  trading (direction required for long/short) and volatility itself,
  which is computed from returns by construction.
\end{itemize}

\paragraph{Data \& Signal.}
\begin{itemize}[leftmargin=1.5em]
  \item \textbf{Financial economics community:} third-party vendor
  (e.g.\ IBES, EODHD): numeric surprise in two forms: percentage
  ($\Delta$EPS\%) and standardised (SUE).%
  \footnote{$\Delta\mathrm{EPS}\%_i = (\mathrm{EPS}_{\mathrm{act}} -
  \mathrm{EPS}_{\mathrm{est}}) / |\mathrm{EPS}_{\mathrm{est}}|$;
  $\mathrm{SUE}_i = (\mathrm{EPS}_{\mathrm{act}} -
  \mathrm{EPS}_{\mathrm{est}}) / \sigma(\varepsilon_i)$.}
  \item \textbf{NLP community:} ECTs, sentiment embeddings capturing
  management tone, credibility, and guidance without explicit numeric
  magnitude.
  \item \textbf{This work (bridge):} earnings news for IE-based numeric
  surprise; ECTs and news for LLM sentiment (\S\ref{sec:dataset}).
\end{itemize}

\paragraph{Entry anchor.}
\begin{itemize}[leftmargin=1.5em]
  \item \textbf{Financial economics community:} enters at next-open, after intraday
  alpha has been priced in and multi-day PEAD has vanished
  \citep{martineau2022rest}, missing the intraday window where
  surprise alpha now concentrates.
  \item \textbf{NLP community:} enters at next-open; qualitative signals
  peak overnight as investors process the earnings narrative.%
  \footnote{ECT transcripts lack sentence-level timestamps, making
  intraday entry infeasible for ECT-based signals.}
  \item \textbf{This work (bridge):} evaluates both intraday and
  next market open horizons, each signal at its natural speed
  (\S\ref{sec:returns}).
\end{itemize}

\paragraph{Trading.}
\begin{itemize}[leftmargin=1.5em]
  \item \textbf{Financial economics community:} dollar-neutral long/short
  : long top-decile SUE (beats), short bottom-decile SUE (misses).
  \item \textbf{NLP community:} no trading framework; predictions are
  evaluated over all announcing stocks, implicitly treating every event
  as equally tradeable.
  \item \textbf{This work (bridge):} dollar-neutral long/short on both
  signal types, following the top-decile/bottom-decile convention
  of the financial economics community.
\end{itemize}

\paragraph{Metric.}
\begin{itemize}[leftmargin=1.5em]
  \item \textbf{Financial economics community:} quintile spread
  (Q5$-$Q1) and cumulative abnormal return (CAR)
  \citep{bernard1989pead}, model-free measures matching canonical
  PEAD reporting.
  \item \textbf{NLP community:} MSE and direction accuracy,
  optimising the low-signal middle of the distribution rather than
  the high-conviction tails where alpha concentrates.
  The failure is structural: NLP evaluation is \textit{pointwise}
  (each stock in isolation) while earnings trading is
  \textit{cross-sectional} (rank all stocks in a session, long top,
  short bottom).
  A model with low MSE can produce a cross-sectional ranking that is
  pure noise.
  \item \textbf{This work (bridge):} IC and Sharpe
  : practitioner-standard metrics from quantitative asset management
  \citep{grinold2000}, complemented by Q5$-$Q1 (\S\ref{sec:eval-metrics}).
\end{itemize}

\section{Full Timestamp Offset Statistics}
\label{sec:appendix-timestamps}

\begin{figure*}[h]
  \centering
  \includegraphics[width=\linewidth]{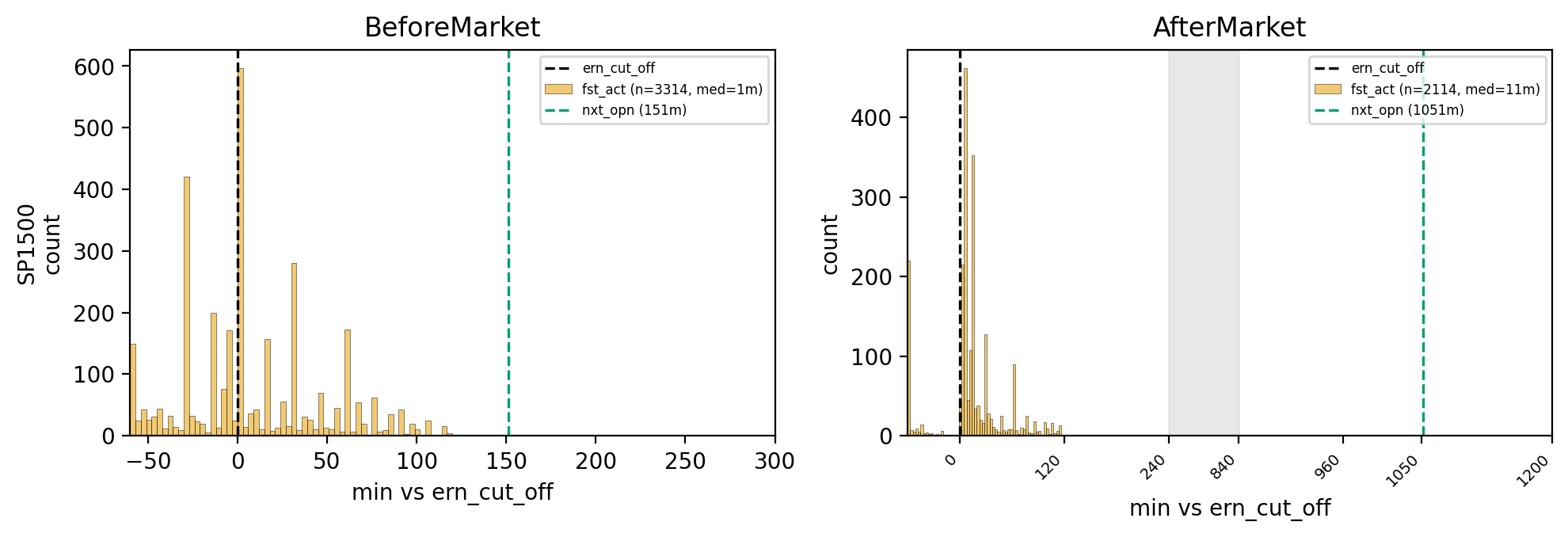}
  \caption{Anchor timestamp offset from earnings cutoff $t_e$ (minutes) 
for BMO (before-market open, $t_e \approx$ 07:00 ET, left) and 
AMC (after-market close, $t_e \approx$ 16:00 ET, right) sessions.
\textit{fst\_act} = \textit{FirstActual} anchor (median 1\,min for BMO, 11\,min for AMC);
\textit{nxt\_opn} = \textit{NextOpen} anchor (09:30 ET open, 151\,min after BMO cutoff, 1051\,min after AMC cutoff).
Gray shading (AMC) marks the overnight window between market close and next open.
BMO news arrives near-simultaneously with the earnings release; 
AMC news arrives within minutes of the release but trades at the next morning's open.}
    \label{fig:timestamp-dist}
\end{figure*}

Figure~\ref{fig:timestamp-dist} shows the empirical distribution of 
\textit{FirstActual} and \textit{NextOpen} anchor offsets relative to 
the earnings cutoff $t_e$, separately for BMO and AMC sessions.

For BMO announcements ($t_e \approx$ 07:00 ET, $n$=3,314), 
\textit{FirstActual} arrives near-simultaneously with the earnings release 
(median offset = 1\,min), well before market open;
\textit{NextOpen} is fixed at 09:30 ET (151\,min after cutoff on average).
The tight clustering around zero confirms that financial news rapidly 
co-reports actuals and estimates at announcement time.

For AMC announcements ($t_e \approx$ 16:00 ET, $n$=2,114),
\textit{FirstActual} arrives within minutes of the release (median = 11\,min),
while \textit{NextOpen} is the following morning's open (median 1,051\,min, $\approx$17.5 hours).
The gray shading marks the overnight window where no trading occurs.
The large gap between \textit{FirstActual} and \textit{NextOpen} for AMC sessions
motivates the intraday entry strategy: alpha is available immediately after the release
but is largely absorbed by next open.

\paragraph{Session definitions.}
A trading session is defined at the (\textit{date}, \textit{before-/after-market}) 
level: all earnings events sharing the same calendar date and market window 
(BMO or AMC) belong to the same session. \textit{FirstActual} is anchored 
to the first news article with confirmed EPS/revenue actuals within that 
session. \textit{NextOpen} is defined as the next regular-session open 
(\texttt{next\_open\_dt}): for AMC announcements, this is 09:30 ET on 
$t_0 + 1$; for BMO announcements on day $t_0$, \textit{NextOpen} 
consolidates both the preceding AMC session (date $t_0 - 1$, after-market) 
and the current BMO session (date $t_0$, before-market), anchoring all 
trades at the 09:30 ET open on $t_0$.

\section{Trading Pipeline and Implementation Details}
\label{sec:appendix-strategies}

\subsection{Embedding Baseline: MLP Pipeline}
\label{sec:appendix-embedding}

The \textit{Embedding} baseline freezes the \href{https://platform.openai.com/docs/guides/embeddings}{text-embedding-3-large} 
(OpenAI, 3072-d) encoder and trains a lightweight MLP on top.

\paragraph{Architecture.}
\begin{verbatim}
Embedding [3072] → StandardScaler
                 → Linear(3072, 256) + ReLU
                 → Linear(256, 128)  + ReLU
                 → Linear(128, 1)
\end{verbatim}
Adam optimizer ($\lambda=10^{-4}$ weight decay), batch size 512, auto-detects CUDA/MPS/CPU.
Two tasks are trained per fold:
\begin{itemize}[leftmargin=1.5em]
  \item \textbf{Classification (sign variant):} BCEWithLogitsLoss; 
  output signal $= P(\text{return} > 0) - 0.5 \in [-0.5, 0.5]$.
  \item \textbf{Regression (magnitude variant):} MSELoss; 
  output signal $=$ predicted return.
\end{itemize}

\paragraph{Rolling-window protocol.}
For each test month $t$:
\begin{enumerate}[leftmargin=1.5em]
  \item \textbf{Split:} train $= [t - T_{\text{train}},\, t - T_{\text{val}} - 1]$; 
  val $= [t - T_{\text{val}},\, t-1]$; test $=$ month $t$.
  \item \textbf{Horizon selection:} train both tasks for each candidate horizon 
  $H \in \{5, 15, 30, 60, 120, 180, 390\}$\,min on pure-train; 
  select $H^* = \arg\max_H \mathrm{IC}_{\text{val}}$ (pooled Spearman rank IC on val).
  \item \textbf{Retrain:} refit selected-horizon models on train $\cup$ val combined.
  \item \textbf{Predict:} append out-of-sample predictions for month $t$.
\end{enumerate}
Final evaluation follows the same IC, Q5$-$Q1, and Sharpe metrics as all other strategies (\S\ref{sec:eval-metrics}).

\subsection{Trading Pipeline: Feature $\to$ Signal $\to$ Weight $\to$ Position}
\label{sec:appendix-pipeline}

\paragraph{Phase 1--2: Raw Features to Standardised Signals.}
EPS surprise:
\[
\mathrm{eps\_surp}_i = \frac{\mathrm{act\_eps}_i - \mathrm{est\_eps}_i}{\mathrm{est\_eps}_i}
\]
Standardised unexpected earnings (SUE):
\[
\mathrm{SUE}_i = \frac{\mathrm{eps\_surp}_i}{\sigma(\varepsilon_i)}
\]
where $\sigma(\varepsilon_i)$ is the rolling standard deviation of historical forecast errors
(window = 8 quarters, min 4), per ticker, shifted one quarter to prevent look-ahead.
Revenue surprise (SUR) is computed analogously.

Note: \textit{FirstActual} strategies use $\Delta\mathrm{EPS}\%$ 
(percentage surprise relative to estimate) as the primary signal,
capturing the immediate market-relevant deviation;
\textit{NextOpen} strategies use SUE (standardised by historical forecast error),
as overnight processing allows incorporating the stock's historical surprise distribution
for better cross-sectional comparability.

\paragraph{Joint signal variants.}
The blended signal $\mathrm{joint}_i = \alpha \cdot \mathrm{SUE}_i + (1-\alpha) 
\cdot \mathrm{SUR}_i$ admits two gating variants, differing only in the 
position entry condition:

\begin{itemize}
    \item \textbf{Joint}: position weight $w_i$ is assigned whenever 
    $|\mathrm{joint\_pctl}_i - 0.5| \geq \tau - 0.5$, using a per-sector 
    expanding-window CDF fitted on the training set.
    
    \item \textbf{Confirmed}: same percentile gate as Joint, with an 
    additional directional agreement filter: a position is opened only 
    when $\mathrm{sign}(\mathrm{SUE}_i) = \mathrm{sign}(\mathrm{SUR}_i)$, 
    requiring EPS and revenue surprises to point in the same direction. 
    This reduces coverage but filters mixed-signal events where EPS and 
    revenue surprises disagree.
\end{itemize}

Both variants use the same $\alpha$ and $\tau$ selected on the validation 
set (2023) and share the same market-wide joint percentile CDF.

\paragraph{Phase 3: Signals to Percentile Ranks.}
An expanding-window empirical CDF fitted market-wide transforms each signal
into a percentile $p_i \in [0,1]$:
\[
p_i = \hat{F}_{\text{market}}(\mathrm{SUE}_i)
\]
The CDF refreshes monthly after the training cutoff.

\paragraph{Phase 4: Percentiles to Weights (Historical Percentile).}
Position weight:
\[
w_i = \frac{G}{N} \times 2 \times (p_i - 0.5) \quad \text{if } |p_i - 0.5| \geq \tau - 0.5
\]
\[
w_i = 0 \quad \text{otherwise}
\]
where $G = 2.0$ (gross target: \$1 long + \$1 short per session),
$N$ = number of events in the session, $\tau = 0.9$ (fixed, top/bottom 10\% trade).
Dollar-neutral by construction.

\paragraph{Phase 4 (PEAD): Cross-Sectional Sort.}
Long leg (top decile):
\[
w_i = +\frac{G/2}{N_{\text{long}}}
\]
Short leg (bottom decile):
\[
w_i = -\frac{G/2}{N_{\text{short}}}
\]
Middle 80\%:
\[
w_i = 0
\]
Sessions with fewer than 10 valid events are skipped.

\paragraph{Hyperparameter Selection.}
Each signal is evaluated at horizons $H \in \{5, 15, 30, 60\}$\,min
for \textit{FirstActual} and $H \in \{5, 15, 30, 60, 120, 180, 390\}$\,min
for \textit{NextOpen}.
The blending weight $\alpha \in \{0.5, 0.6, 0.7, 0.8, 0.9\}$ is selected
on the validation set (2023) by maximising portfolio Sharpe ratio, subject to per-session IC $> 0$ to ensure positive signal direction,
and held fixed for test (2024--2025); $\tau{=}0.9$ is fixed throughout.

\paragraph{Phase 5--6: Execution.}
Portfolio PnL per session:
\[
\mathrm{PnL}_d = \sum_i
w_i \times r_{i,H}
\]
where $r_{i,H}$ is the forward return at horizon $H \in \{5,15,30,60,120,180,390\}$\,min.
Risk controls (sector caps, exposure limits) are disabled by default.
\section{Evaluation Metric Definitions}
\label{sec:appendix-metrics}

\begin{description}[leftmargin=1.5em, labelwidth=!, labelsep=0.5em,
  font=\normalfont\bfseries]

  \item[IC (Information Coefficient).] Spearman rank correlation between
  the signal and realized forward returns, measuring signal quality
  independently of position sizing.
  
  \textbf{Per-session IC} (primary metric): computed cross-sectionally 
  within each event date $d$:
  \[
  \mathrm{IC}_d = \mathrm{Spearman}\bigl(\mathbf{s}_d,\, \mathbf{r}_d\bigr)
  \]
  where $\mathbf{s}_d$ = signal scores and $\mathbf{r}_d$ = forward returns 
  for all stocks announcing on date $d$.
  The reported IC is the time-average $\mathrm{E}[\mathrm{IC}_d]$ across all sessions.

  \textbf{Pooled IC}: computed across all events simultaneously (used in 
  Figure~\ref{fig:return-ic} for characterising signal decay); 
  typically higher than per-session IC due to cross-sectional variation in base returns.
\textbf{$t_{\mathrm{NW}}$ (Newey-West $t$-statistic)}: tests whether 
$\mathrm{E}[\mathrm{IC}_d] \neq 0$, correcting for serial correlation 
across sessions:
\begin{equation}
    t_{\mathrm{NW}} = \frac{\bar{\mathrm{IC}}}{\hat{\sigma}_{\mathrm{NW}} / \sqrt{T}}
\end{equation}
where $\bar{\mathrm{IC}}$ is the time-averaged IC, $T$ is the number of 
sessions, and $\hat{\sigma}_{\mathrm{NW}}$ is the Newey-West standard 
deviation correcting for serial correlation up to lag $L$ (we use $L=5$). 
$|t_{\mathrm{NW}}| > 2$ corresponds to $p < 0.05$.

  \item[Q5$-$Q1 bp ($t$).] Fama-MacBeth session-averaged quintile spread:
  \[
  \mathrm{Q5{-}Q1}_d = \bar{r}_{d,Q5} - \bar{r}_{d,Q1}
  \]
  where $\bar{r}_{d,Q5}$ and $\bar{r}_{d,Q1}$ are mean returns of the 
  top and bottom quintile on date $d$, expressed in basis points (bp).
  The $t$-statistic uses Fama-MacBeth standard errors \citep{bernard1989pead}.
  Measures tail separability independent of the full cross-section.

  \item[Sharpe Ratio.]
  \[
  \mathrm{Sharpe} = \frac{\bar{R}_p - R_f}{\sigma_p}
  \]
  where $\bar{R}_p$ is mean annualised portfolio return, $R_f$ the risk-free
  rate (set to zero for simplicity), and $\sigma_p$ the annualised return 
  standard deviation. Measures overall strategy quality by incorporating 
  signal quality, selectivity, and sizing.

  \item[Tot.\ Ret.] Cumulative total return over the test period (2024--2025),
  expressed as a percentage. Under dollar-neutral long/short sizing,
  equivalent to market-adjusted cumulative abnormal return (CAR).

  \item[$N$.] Number of trades (long $+$ short positions) over the test period.

\end{description}

\section{Extraction Validation: News vs.\ EOD Vendor}
\label{sec:appendix-validation}

\begin{figure*}[h]
  \centering
  \includegraphics[width=\linewidth]{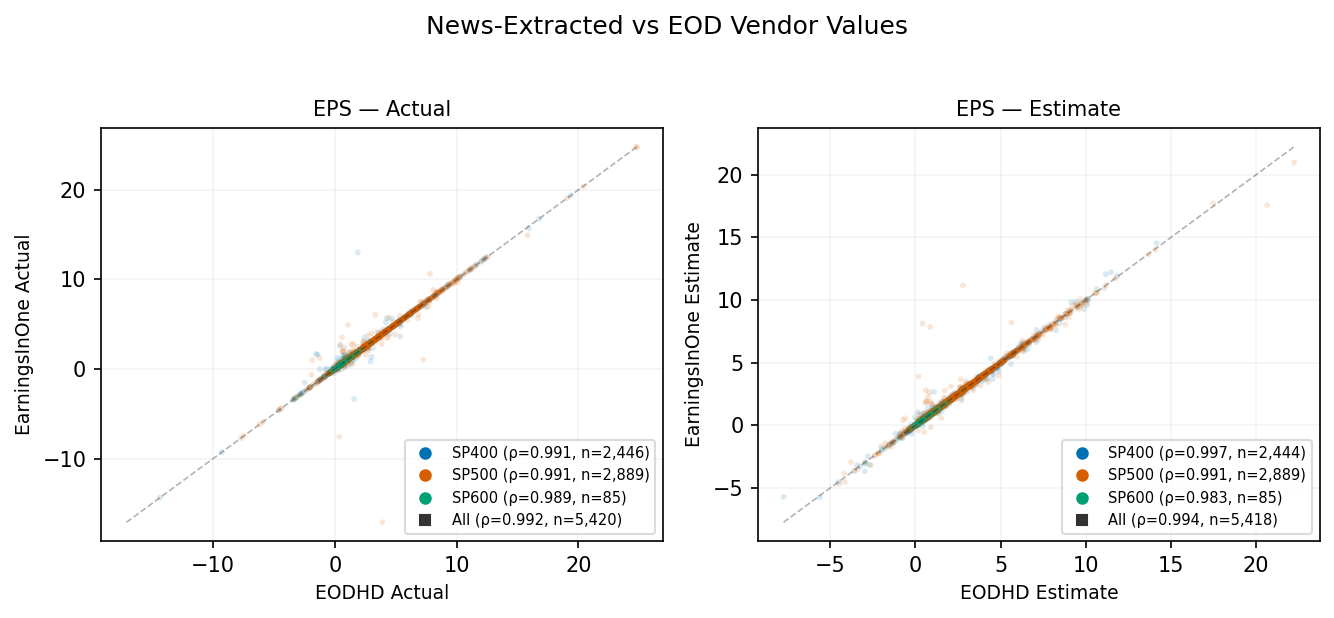}
 \caption{Scatter plots of \datasetname{} LLM-  extracted values vs.\ EODHD vendor values
    for EPS actual (left) and EPS estimate (right), by index.
    Spearman $\rho \geq 0.989$ for actuals and $\rho \geq 0.983$ for estimates across all indices,
    confirming near-perfect agreement. Dashed line = identity.}
  \label{fig:eod-scatter}
\end{figure*}

\begin{figure}[h]
  \centering
  \includegraphics[width=\linewidth]{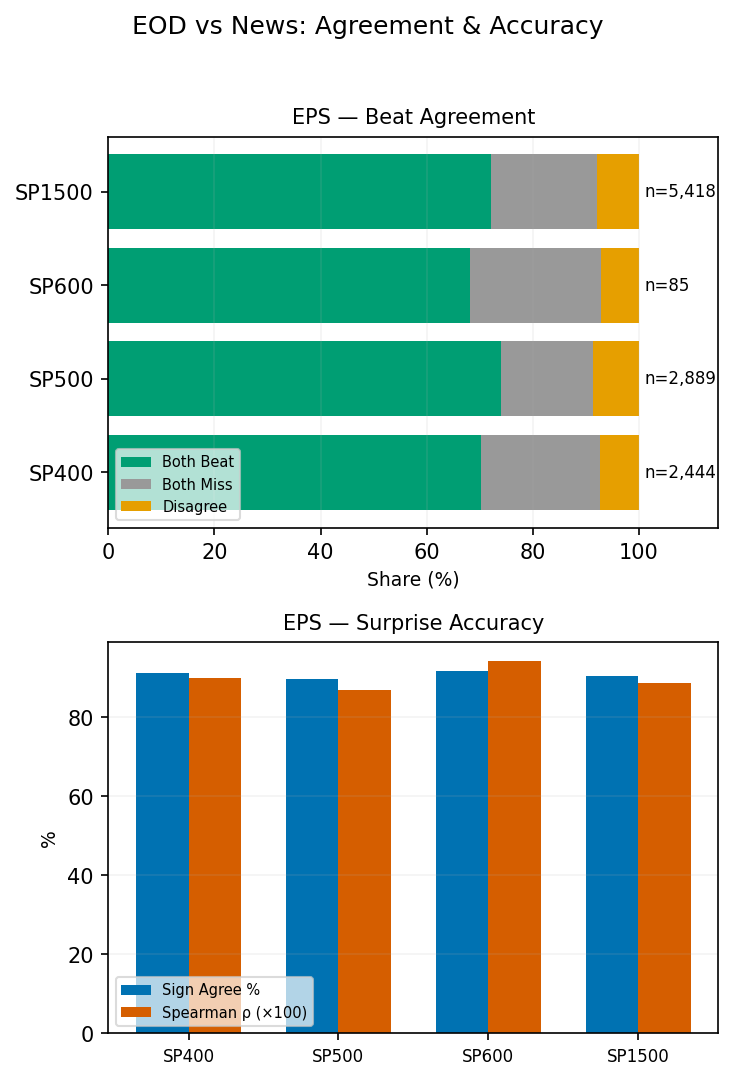}
 \caption{Beat/miss agreement and surprise accuracy between \datasetname{} and EODHD.
\textbf{Top:} stacked bars show the share of both-beat, both-miss, and disagreement events;
overall agreement exceeds 90\% across all indices.
\textbf{Bottom:} sign agreement rate (\%) and Spearman $\rho$ ($\times 100$) by index,
both consistently above 85\%, confirming reliable directional extraction.}
  \label{fig:eod-agreement}
\end{figure}

Figures~\ref{fig:eod-scatter} and~\ref{fig:eod-agreement} compare
LLM-extracted EPS and revenue values from \datasetname{} against
ground-truth values from an EOD financial data vendor, across all four
equity universes.

\section{Dataset Construction Details}
\label{sec:appendix-dataset}

\paragraph{News Collection}
News articles are sourced from a licensed financial wire aggregator.
We filter to earnings-related articles by requiring at least one of:
(a) ticker mention in headline or lede matching the announcing company,
(b) keyword match on earnings-related terms (``earnings'', ``EPS'', ``quarterly results'')
Duplicate articles (same headline within a 10-minute window) are
deduplicated by URL hash; the collection window spans $[-$60,min, +120,min] around each 8-K filing timestamp.

\paragraph{Privacy and Content.}
Financial news articles and earnings conference call transcripts may contain
names of corporate executives and financial analysts acting in their
professional capacity; these are already part of the public record and
do not constitute private personally identifiable information (PII).
No private individual data is collected or released.
To further limit exposure, we release article IDs and extracted metadata
only, not full article text.
No offensive content was identified in the corpus; all sources are
regulated financial media outlets subject to editorial standards.

\paragraph{Dataset Documentation.}
\datasetname{} covers English-language U.S.\ financial news and earnings conference
call transcripts for SP\,1500 companies (2022--2025).
The corpus is domain-specific (financial reporting) and monolingual (English).
Company coverage skews toward large- and mid-cap firms (SP\,500 and SP\,400 represent
98\% of events); SP\,600 small-cap firms are underrepresented due to lower news coverage.
No individual-level demographic data is present; all entities are publicly listed
U.S.\ corporations subject to SEC reporting requirements.
Linguistic phenomena include financial jargon, numerical expressions,
forward-looking statements, and structured analyst Q\&A dialogue.

\paragraph{Data Consent.}
All data in \datasetname{} is sourced from publicly available corporate disclosures
and licensed commercial APIs.
Earnings conference call transcripts are publicly disclosed under SEC Regulation FD,
which mandates that material information be made available to all investors simultaneously.
Financial news articles and price data are accessed via the EODHD Historical Data API
under a commercial license agreement.
No individual consent is required as all sources constitute public corporate records.

\paragraph{Ethics Review.}
This work does not involve human subjects research, collection of personal data,
or any procedures requiring ethics board review.
All data sources are publicly available corporate disclosures or licensed
commercial data products.

\paragraph{Dataset Statistics}

Table~\ref{tab:split_dist} reports event and ticker counts by index and temporal split.
The corpus totals 5{,}428 earnings events across 701 unique tickers over 2022--2025.
SP\,500 and SP\,400 dominate (2,897 and 2,446 events respectively),
while SP\,600 contributes only 85 events ($\sim$1.6\% of the corpus),
reflecting the lower financial news coverage of small-cap firms.
The test set (2024--2025) is the largest split by design,
covering 2,468 events across 640 unique tickers.
Ticker counts across splits sum to more than the ``All'' column
because some tickers report in multiple years.

\begin{table*}
\centering
\small
\resizebox{\textwidth}{!}
{\begin{tabular}{lrrrrrrrr}
\toprule
 & \multicolumn{4}{c}{Earnings} & \multicolumn{4}{c}{Tickers} \\
 & Train (2022) & Val (2023) & Test (2024-25) & All & Train (2022) & Val (2023) & Test (2024-25) & All \\
\midrule
SP400 & 768 & 589 & 1089 & 2446 & 277 & 252 & 304 & 354 \\
SP500 & 810 & 766 & 1321 & 2897 & 285 & 280 & 319 & 346 \\
SP600 & 11 & 16 & 58 & 85 & 5 & 12 & 29 & 31 \\
SP1500 & 1589 & 1371 & 2468 & 5428 & 565 & 541 & 640 & 701 \\
\bottomrule
\end{tabular}}

\caption{EarningsInOne dataset statistics by index and split.
Earnings = number of earnings events; Tickers = number of unique tickers.
SP 600 (small-cap) is underrepresented; small-cap firms attract less financial news coverage, consistent with their lower analyst coverage and trading volume.
Strict temporal split: train 2022, validation 2023, test 2024--2025.}
\label{tab:split_dist}
\end{table*}

\section{Prompt Templates}
\label{sec:appendix-prompt}
Full prompt templates for IE extraction and sentiment scoring are
available in the released code repository. IE extraction uses a
structured JSON output schema with chain-of-thought reasoning;
sentiment scoring uses a single-pass instruction with a
$[-1,+1]$ scale and verbal anchors at $\pm 0.5$ and $\pm 1.0$.

\paragraph{ECT Sentiment Scoring Prompt.}

\begin{tcolorbox}[
  colback=gray!8, 
  colframe=gray!40, 
  arc=3pt, 
  boxrule=0.4pt,
  breakable,
  enhanced
]

\small
You are a financial analyst predicting how the market will react to this earnings call.
You will be given the full transcript of a company's quarterly earnings call. \\

\textbf{Your task:} assess whether the call conveys \emph{incrementally} positive or negative
information --- i.e., things that would surprise the market beyond what was already
expected from the headline numbers (EPS/revenue beat or miss).

Focus on (in order of importance):
\begin{enumerate}[leftmargin=1.5em]
  \item \textbf{Q\&A section} --- analyst tone (hostile vs.\ friendly), management evasiveness
  or defensiveness, unexpected admissions, stumbling on questions
  \item \textbf{Forward guidance} --- specific vs.\ vague, raised/maintained/lowered,
  hedging language (``we hope'', ``we'll try'' vs.\ ``we will'', ``we expect'')
  \item \textbf{Tone shifts} --- prepared remarks upbeat but Q\&A defensive = bearish signal
  \item \textbf{Incremental disclosure} --- new risks mentioned, customer losses,
  pipeline changes, margin pressure, one-time items framed as recurring
\end{enumerate}

\textbf{Scoring guidance} (use the full range):
\begin{itemize}[leftmargin=1.5em]
  \item sentiment near $+1.0$: management clearly signalling acceleration,
  raising guidance with specifics, comfortable Q\&A, analysts impressed
  \item sentiment near $+0.3$: slightly better than expected tone, modest positive
  guidance, no red flags
  \item sentiment near $0.0$: neutral, in-line, no surprises either direction
  \item sentiment near $-0.3$: slight hedging, vague guidance, some analyst pushback
  \item sentiment near $-1.0$: defensive Q\&A, guidance cut, evasive on key metrics,
  analyst frustration visible
\end{itemize}

Respond with a JSON object:
\begin{verbatim}
{ "sentiment": <float between -1.0 and 1.0> }
\end{verbatim}

\end{tcolorbox}

\medskip

\paragraph{ECT Decomposition Scoring Prompt.}
\begin{tcolorbox}[colback=gray!8, colframe=gray!40, arc=3pt, boxrule=0.4pt,  breakable]

\small
You are a financial analyst predicting how the market will react to this earnings call.
You will be given the full transcript of a company's quarterly earnings conference call. \\

\textbf{Your task:} score each of the following dimensions independently,
based \emph{only} on what is discussed in the call. Each dimension captures
a distinct aspect of the call's information content --- score them as
independently as possible.

Focus on (in order of importance):
\begin{enumerate}[leftmargin=1.5em]
  \item \textbf{Q\&A section} --- analyst tone (hostile vs.\ friendly), management evasiveness
  or defensiveness, unexpected admissions, stumbling on questions
  \item \textbf{Forward guidance} --- specific vs.\ vague, raised/maintained/lowered,
  hedging language (``we hope'', ``we'll try'' vs.\ ``we will'', ``we expect'')
  \item \textbf{Tone shifts} --- prepared remarks upbeat but Q\&A defensive = bearish signal
  \item \textbf{Incremental disclosure} --- new risks mentioned, customer losses,
  pipeline changes, margin pressure, one-time items framed as recurring
\end{enumerate}

\textbf{Scoring guidance} (use the full range):
\begin{itemize}[leftmargin=1.5em]
  \item \textbf{Directional dimensions}
  (\texttt{eps\_beat\_framing}, \texttt{rev\_quality}, \texttt{margin\_trajectory},
  \texttt{cash\_flow\_quality}, \texttt{guidance\_direction}, \texttt{long\_term\_outlook},
  \texttt{demand\_momentum}, \texttt{pricing\_power}, \texttt{market\_share},
  \texttt{investment\_cycle}, \texttt{competitive\_threat}, \texttt{regulatory\_risk},
  \texttt{supply\_chain\_health}, \texttt{prepared\_vs\_qa\_gap},
  \texttt{analyst\_tone}, \texttt{analyst\_surprise}):
  near $+1.0$ = strongly positive for the stock;
  near $0.0$ = neutral/in-line;
  near $-1.0$ = strongly negative for the stock
  \item \textbf{Intensity dimensions}
  (\texttt{guidance\_specificity}, \texttt{guidance\_conviction},
  \texttt{macro\_sensitivity}, \texttt{mgmt\_confidence}, \texttt{transparency},
  \texttt{surprise\_novelty}, \texttt{qa\_depth}):
  near $1.0$ = strong presence; near $0.0$ = absent/minimal
  \item Use \texttt{null} for any dimension not assessable from the transcript
\end{itemize}

Respond with a JSON object:
{\footnotesize
\begin{verbatim}
{ "eps_beat_framing": 
  <float -1.0 to 1.0 or null>,
  "rev_quality":    
  <float -1.0 to 1.0 or null>,
  "margin_trajectory":  
  <float -1.0 to 1.0 or null>,
  "cash_flow_quality":  
  <float -1.0 to 1.0 or null>,
  "guidance_direction": 
  <float -1.0 to 1.0 or null>,
  "guidance_specificity":
  <float  0.0 to 1.0 or null>,
  "guidance_conviction":
  <float  0.0 to 1.0 or null>,
  "long_term_outlook":  
  <float -1.0 to 1.0 or null>,
  "demand_momentum":    
  <float -1.0 to 1.0 or null>,
  "pricing_power":      
  <float -1.0 to 1.0 or null>,
  "market_share":       
  <float -1.0 to 1.0 or null>,
  "investment_cycle":   
  <float -1.0 to 1.0 or null>,
  "competitive_threat": 
  <float -1.0 to 1.0 or null>,
  "regulatory_risk":    
  <float -1.0 to 1.0 or null>,
  "supply_chain_health":
  <float -1.0 to 1.0 or null>,
  "macro_sensitivity":  
  <float  0.0 to 1.0 or null>,
  "mgmt_confidence":    
  <float  0.0 to 1.0 or null>,
  "transparency":       
  <float  0.0 to 1.0 or null>,
  "prepared_vs_qa_gap": 
  <float -1.0 to 1.0 or null>,
  "surprise_novelty":  
  <float  0.0 to 1.0 or null>,
  "analyst_tone":       
  <float -1.0 to 1.0 or null>,
  "analyst_surprise":   
  <float -1.0 to 1.0 or null>,
  "qa_depth":           
  <float  0.0 to 1.0 or null> }
\end{verbatim}
}
\end{tcolorbox}

\section{Full Results Table}
\label{sec:appendix-full-results}

Table~\ref{tab:full-results} reports all quantitative and qualitative
variants, including those omitted from the main table for concision.
The table is organised into four groups.
\textbf{Na\"{i}ve baselines} provide no-signal floors: passive
index benchmarks (SPY, QQQ), random long/short, long-only, and
two prior-NLP baselines (Ticker-Mean and ECT Embedding).
\textbf{Quantitative (Fast)} variants isolate four factors:
ranking mechanism (PEAD cross-sectional sort vs.\ historical
percentile), entry anchor (\textit{FirstActual} vs.\
\textit{NextOpen}), source breadth (EPS-only vs.\ Joint
EPS+revenue), and data quality (LLM-extracted EN vs.\ vendor EOD).
\textbf{Qualitative (Slow)} variants cover ECT sentiment under
multiple ablations (number masking, numeric grounding, alternative
LLM backends, prepared-remarks-only, Q\&A-only) and news sentiment
aggregated to \textit{NextOpen}.
\textbf{Combined (Combo)} variants directly merge ECT sentiment
with quantitative signals at \textit{NextOpen}, isolating the
effect of signal combination at a shared entry anchor.
\begin{table*}[h]
\centering
\small

\resizebox{\textwidth}{!}{%
\begin{tabular}{ll l ccccc}
\toprule
\textbf{Category} & \textbf{Strategy} & \textbf{Entry anchor} & IC ($t_{\mathrm{NW}}$) & Q5$-$Q1 bp ($t$) & Sharpe & Tot.\ Ret. & $N$ \\
\midrule
\multirow{5}{*}{Na\"{i}ve}
 & Random L/S             & NextOpen    & ---                   & ---               & $-$0.88 & $-$10.5\% & 2426 \\
 & Long-only     & NextOpen    & ---                   & ---               & 0.11    & 5.8\%     & 2077 \\
 & \textit{Ticker-Mean}   & NextOpen    & ---                   & ---               & $-$3.13 & $-$2.2\%  & 2360 \\
 & \textit{Embedding (sign)}  & NextOpen    & ---                   & ---               & 0.80    & 23.3\%   & 2468 \\
 & \textit{Embedding (mag)}   & NextOpen    & ---                   & ---               & 0.56    & 13.9\%   & 2468 \\
\midrule
\multirow{7}{*}{\shortstack{Quantitative\\(Fast)}}
 & \textit{PEAD-EPS(EN)}   & NextOpen    & 0.03 (0.97)           & 33.9 (1.36)       & 3.31    & 29.1\%  & 380 \\
 & \textit{PEAD-Joint(EN)} & NextOpen    & 0.03 (0.84)            & 42.1 (1.60) & 3.18   & 25.2\%  & 326 \\
 & \textit{EPS(EN)}        & NextOpen    & $-$0.02 ($-$0.56)     & $-$5.4 ($-$0.12)  & $-$0.31 & $-$4.2\%  & 400 \\
 & \textit{Joint(EN)}      & NextOpen    & $-$0.02 ($-$1.03)   & 19.87 (0.47)     & 0.23    & 6.3\%      & 242 \\
 & \textit{EPS(EN)}        & FirstActual & 0.03 (0.94)           & \textbf{137.7 (3.54)}      & 1.81    & 50.4\%  & 400 \\
 & \textit{Joint(EN)}      & FirstActual & 0.05 (1.58)           & 61.3 (1.94)       & 1.79    & 39.9\%  & 296 \\
 & \textit{Joint(EOD)}$^\dagger$ & FirstActual & \textbf{0.08 (2.40)}   & 118.2 (4.13) & \textbf{2.43} & \textbf{64.5\%}  & 343 \\
\midrule
\multirow{11}{*}{\shortstack{Qualitative\\(Slow)}}
 & \textit{Sentiment(ECT)}          & NextOpen    & \textbf{0.11 (4.04)} & {153.0 (4.30)} & 2.28 & 63.6\%  & 591 \\
 & \textit{Sentiment(ECT-mask)}     & NextOpen    & 0.07 (2.62) & 118.7 (3.96) & 0.40  & 7.0\%   & 412 \\
 & \textit{Sentiment(ECT-informed)} & NextOpen    & 0.06 (2.10) & 94.1 (2.62)  & \textbf{2.65}  & \textbf{72.0\%}  & 599 \\
 & \textit{Sentiment(ECT-Haiku)}    & NextOpen    & 0.08 (2.43) & \textbf{213.2 (4.01)} & 1.55  & 31.4\%  & 455\\
 & \textit{Sentiment(ECT-Gemini)}   & NextOpen    & 0.08 (2.77) & 71.1 (2.01) & 1.16 & 12.6\%   & 110 \\
 & \textit{Sentiment(EN-agg)}       & NextOpen    & 0.05 (1.89) & $-$4.6 ($-$0.24) & 0.42  & 5.9\%   & 496 \\

 & \textit{Sentiment(ECT-Prepared)}  & NextOpen    & 0.05 (1.69) & 31.5 (0.98) & 2.15 & 40.8\%  & 430 \\
 & \textit{Sentiment(ECT-QA)}        & NextOpen    & 0.07 (2.42) & 76.6 (2.83) & 1.47 & 27.9\%  & 492 \\

   & \textit{PEAD-Sentiment(ECT)}     & NextOpen    & 0.08 (2.70) & 136.4 (3.94) & $-$0.60 & $-$8.1\%  & 578\\
 & \textit{PEAD-Sentiment(ECT-mask)} & NextOpen   & 0.04 (1.25) & 18.9 (1.21) & $-$0.92 & $-$2.2\%  & 568  \\
 & \textit{PEAD-Sentiment(ECT-informed)} & NextOpen & 0.09 (2.76) & 41.2 (1.77) & 1.63 & 11.5\%  & 590  \\
 & \textit{PEAD-Sentiment(ECT-Haiku)} & NextOpen   & 0.0713 (2.59) & 138.3 (4.23) & -0.15 & -2.0\% & 621  \\
 & \textit{PEAD-Sentiment(ECT-Gemini)} & NextOpen & 0.0603 (2.22) & 20.7 (1.06) & 0.77 & 9.5\% & 561 \\

\midrule
\multirow{2}{*}{\shortstack{Combined\\(Combo)}}
 & \textit{Sentiment(ECT)+Joint(EN)} & NextOpen    & 0.05 (1.99) & 20.3 (0.79) & 0.59 & 6.3\% & 271 \\
 & \textit{Sentiment(ECT)+EPS(EN)}   & NextOpen    & 0.06 (2.37) & 86.3 (2.00) & 0.12 & 2.1\% & 437 \\

\bottomrule
\end{tabular}%
}

\caption{Out-of-sample trading performance, test set (2024--2025 with 2{,}468 events), SP\,1500.
\textbf{Entry anchors:}
\textit{FirstActual} = first news article with confirmed EPS/revenue actuals;
\textit{NextOpen} = 09:30\,am on the following trading day.
\textbf{Metrics:}
IC\,($t_{\mathrm{NW}}$) = Spearman rank IC \citep{grinold2000} computed per event date
($t_{\mathrm{NW}}$: Newey-West $t$-statistic correcting for serial correlation);
Q5$-$Q1\,bp\,($t$) = Fama-MacBeth date-averaged quintile spread \citep{bernard1989pead}
($t$: Fama-MacBeth standard errors);
$|t|>2$ corresponds to $p<0.05$ for both;
$N$ = number of trades.
\textbf{Signals:}
\textit{Joint} = EPS + revenue surprise combined;
$^\dagger$EOD: EODHD Historical Data pre-release estimates; professional-grade vendor actuals with superior data quality, but not available in real time for live deployment;
\textit{Embedding} (sign/mag variants, text-embedding-3-large, OpenAI) = ECT embeddings passed through a 2-layer MLP;
\textit{PEAD-Sentiment(ECT)} variants = ECT sentiment strategies using canonical PEAD contemporaneous cross-sectional sort (vs.\ historical percentile for vanilla ECT);
\textit{Sentiment(ECT-Prepared)} = sentiment scored over prepared remarks only;
\textit{Sentiment(ECT-QA)} = sentiment scored over Q\&A section only;
\textit{Sentiment(ECT-Haiku/Gemini)} = ECT sentiment using alternative LLM backends;
\textit{Combo} = direct merger of \textit{Sentiment(ECT)} with quantitative signal at \textit{NextOpen}.
All other variants use historical percentile calibration unless labelled PEAD.s
\textbf{Metric Interpretation:}
IC measures overall signal quality (does the signal rank stocks correctly?);
Q5$-$Q1 measures tail separability (do the extreme quintiles trade well?);
Sharpe measures full strategy performance (does it make money risk-adjusted?).
High IC with low Sharpe indicates the signal ranks stocks well but individual trade returns are noisy and volatile;
high Sharpe with low IC signals selectivity-driven returns (strong performance concentrated in a small number of high-conviction trades).
\textbf{Bold} = best per column within group.
\textbf{Passive Market benchmarks:} SPY and QQQ are the most widely traded ETFs tracking the S\&P\,500 and Nasdaq-100, serving as buy-and-hold baselines: SPY total return 48\% (Sharpe 1.02); QQQ total return 54\% (Sharpe 0.94).}
\label{tab:full-results}

\end{table*}

\FloatBarrier
\begin{table}
\centering
\small
\caption{Per-dimension IC at 390\,min (\textit{NextOpen}) for the
23 ECT decomposition dimensions, SP\,1500 (full dataset, 2022--2025).
Ordered by IC descending within each category.
Positive = directional signal toward next-open return;
negative = contrarian or noise.}
\label{tab:decompose-ic}
\begin{tabular}{llc}
\toprule
\textbf{Category} & \textbf{Dimension} & \textbf{IC} \\
\midrule
\multirow{4}{*}{\shortstack{A. Financial\\Performance}}
  & Cash flow quality      & 0.055 \\
  & Margin trajectory      & 0.022 \\
  & EPS beat framing       & 0.025 \\
  & Revenue quality        & 0.018 \\
\midrule
\multirow{4}{*}{\shortstack{B. Forward \\Guidance} }
  & Guidance direction     & 0.020 \\
  & Guidance specificity   & 0.012 \\
  & Guidance conviction    & 0.008 \\
  & Long-term outlook      & 0.007 \\
\midrule
\multirow{4}{*}{\shortstack{C. Growth \& \\Demand}}
  & Market share           & 0.028 \\
  & Pricing power          & 0.023 \\
  & Demand momentum        & 0.001 \\
  & Investment cycle       & $-$0.003 \\
\midrule
\multirow{4}{*}{\shortstack{D. Risk \& \\ External}}
  & Competitive threat     & 0.041 \\
  & Regulatory risk        & 0.021 \\
  & Supply chain health    & 0.020 \\
  & Macro sensitivity      & $-$0.007 \\
\midrule
\multirow{4}{*}{\shortstack{E. Management \\Communication}}
  & Mgmt confidence        & 0.021 \\
  & Transparency           & 0.014 \\
  & Prepared vs.\ Q\&A gap & 0.004 \\
  & Surprise novelty       & $-$0.029 \\
\midrule
\multirow{3}{*}{\shortstack{F. Analyst \\Reception}}
  & Analyst surprise       & 0.038 \\
  & Analyst tone           & 0.037 \\
  & Q\&A depth             & $-$0.009 \\
\bottomrule
\end{tabular}
\end{table}

\section{ECT Decomposition: Per-Dimension IC}
\label{sec:appendix-decompose}

Table~\ref{tab:decompose-ic} reports Spearman IC at 390\,min
(\textit{NextOpen}) for each of the 23 orthogonal ECT dimensions,
computed on the full dataset (2022--2025).
Dimensions are scored independently by a structured LLM prompt
(Appendix~\ref{sec:appendix-prompt}); see \S\ref{sec:exp-analysis}
for interpretation.

The 23 dimensions span six categories:
\textbf{A. Financial Performance} (backward-looking): EPS beat framing,
revenue quality, margin trajectory, cash flow quality --- capturing how
management characterises reported results and whether earnings quality
is sustainable.
\textbf{B. Forward Guidance}: guidance direction
(raised/maintained/lowered), specificity, conviction, long-term outlook
--- measuring the strength and credibility of management's forward
outlook.
\textbf{C. Growth \& Demand}: demand momentum, pricing power, market
share, investment cycle --- reflecting whether the business is
accelerating or decelerating in its competitive environment.
\textbf{D. Risk \& External}: competitive threat, regulatory risk,
supply chain health, macro sensitivity --- assessing exposure to
external forces beyond management control.
\textbf{E. Management Communication Quality}: overall confidence,
transparency, prepared-vs-Q\&A tone gap, surprise novelty --- evaluating
how clearly and honestly management communicates, and whether new
information is disclosed.
\textbf{F. Analyst Reception}: analyst tone, analyst surprise, Q\&A
depth --- capturing real-time expert reaction as a proxy for
institutional investor sentiment.
Each dimension is scored on $[-1, +1]$ independently, capturing
incremental information beyond the headline EPS/revenue outcome.

\end{document}